\documentclass[runningheads,a4paper]{llncs}

\usepackage{amssymb}
\usepackage{graphicx}
\usepackage{subfigure}
\usepackage{booktabs}
\usepackage{multirow}
\usepackage{color}

\usepackage{url}
\urldef{\mailsb}\path|{eduardo.aguilar,marc.bolanos,petia.ivanova}@ub.edu | 
\newcommand{\keywords}[1]{\par\addvspace\baselineskip
\noindent\keywordname\enspace\ignorespaces#1}

\begin{document}

\mainmatter  

\title{Exploring  Food Detection using CNNs}

\titlerunning{Exploring Food Detection using CNNs}

%
%
\author{Eduardo Aguilar
%
\and Marc Bola\~nos
\and Petia Radeva
}
%

\institute{Universitat de Barcelona and Computer Vision Center, Spain
\\
\mailsb\\
}
%
%

\toctitle{Lecture Notes in Computer Science}
\tocauthor{Authors' Instructions}
\maketitle

\begin{abstract}
One of the most common critical factors directly related to the cause of a chronic disease is unhealthy diet consumption. In this sense, building an automatic system for food analysis could allow a better understanding of the nutritional information with respect to the food eaten and thus it could help in taking corrective actions in order to consume a better diet. The Computer Vision community has focused its efforts on several areas involved in the visual food analysis such as: food detection, food recognition, food localization, portion estimation, among others. For food detection, the best results evidenced in the state of the art were obtained using Convolutional Neural Network. However, the results of all these different approaches were gotten on different datasets and therefore are not directly comparable. This article proposes an overview of the last advances on food detection and an optimal model based on GoogLeNet Convolutional Neural Network method, principal component analysis, and a support vector machine that outperforms the state of the art on two public food/non-food datasets.
\keywords{CNN, PCA, GoogLeNet, SVM, Food Detection.}

\end{abstract}
\section{Introduction}
In the last decades, the amount of people with overweight and obesity is progressively increasing \cite{ng2014}, whom generally maintain an excessive unhealthy diet consumption. Additionally to the physical and psychological consequences involved to their condition, these people are more prone to acquire chronic diseases such as heart disease, respiratory diseases, and cancer \cite{who2003}. Consequently, it is highly necessary to build tools that offer high accuracy in nutritional information estimation from ingested foods, and thus, improve the control of food consumption and treat people with nutritional problems. 

Recently, the computer vision community has focused its efforts on several areas involved on developing automated systems for visual food analysis, which usually involves a food detection method \cite{kagaya2014,bolanos2016,myers2015,Singla16}. This method, also called as food/non-food classification, has as purpose to determine the presence or absence of food in an image. Generally it is applied as pre-processing of images prior to visual food analysis, and also can be useful for cleaning a huge amount of images acquired from the WEB or wearable cameras, leaving only images that contain food. 

The food detection has been  investigated in the literature in different works  \cite{kagaya2014,Singla16,kitamura2009,farinella2015,ragusa2016}, where has been evidenced that the best results obtained are based on Convolutional Neural Networks (CNN). The first method based on CNN for food detection was proposed by \cite{kagaya2014}, that achieved a 93.8\% using AlexNet model \cite{krizhevsky2012} on a dataset composed of 1234 food images and 1980 non-food images acquired from social media sources, which implies a 4\% higher than accuracy with respect to the hand crafted features \cite{kitamura2009}. In
\cite{kagaya2015}, the authors improved the accuracy on this dataset to 99.1\% using the NIN model \cite{lin2013}. In addition, they evaluated their model on other datasets, IFD and FCD, obtaining 95\% and 96\% of accuracy, respectively. An evaluation on a huge dataset with over 200,000 images constructed from Food101 \cite{food101} and ImageNet Challenge was done in \cite{myers2015}, the authors achieved 99.02\% using an efficient CNN model based on inception module called GoogLeNet \cite{szegedy2015}.  The same model was used in \cite{bolanos2016}, the authors obtained 95.64\% of accuracy to evaluate GoogLeNet on a dataset composed of Food101; food-related images extracted from the ImageNet Challenge dataset; and Pascal \cite{everingham2010} (used as non-food images). An evaluation of different CNN models and settings was proposed by \cite{ragusa2016} on a dataset, we call RagusaDS. The authors got the best results using AlexNet for feature extraction and Binary SVM \cite{cortes1995} for classification. In terms of accuracy, they achieved 94.86\%. In \cite{Singla16}, the authors apply fine-tuning on the last six layers of a GoogLeNet obtaining high accuracy, but tested their model on only 5000 image (Food-5k) which contains the same amount of food and non-food images. 

Since the proposed models were evaluated on different datasets, the results obtained are not directly comparable. Therefore, from the review of the literature, we selected the public datasets with more than 15,000 images. Based on this requirement, we used RagusaDS from \cite{ragusa2016} and FCD \cite{kagaya2015} datasets for the food detection analysis.


Furthermore, we explored the food detection using the GoogLeNet, because this CNN model has presented the best results in the classification of objects, and in particular for food detection it has also presented good results on multiplies datasets with images acquired in different conditions \cite{bolanos2016,myers2015,Singla16}. Specifically, we propose a food detection model based on GoogLeNet for feature extraction, PCA \cite{jolliffe2002} for feature selection and SVM for classification that proves the best accuracy in the state of the art with respect to the previous works on the same datasets. 


The organization of this article is as follows. In section II, we  present a brief description of the datasets used and the proposed methodology. In section III, we  present and discuss the results obtained. Finally in section IV, we describe the drawn conclusions and future work.
\newpage

\section{Methodology for food detection}

\begin{figure}
\centering
\includegraphics[width=1\textwidth]{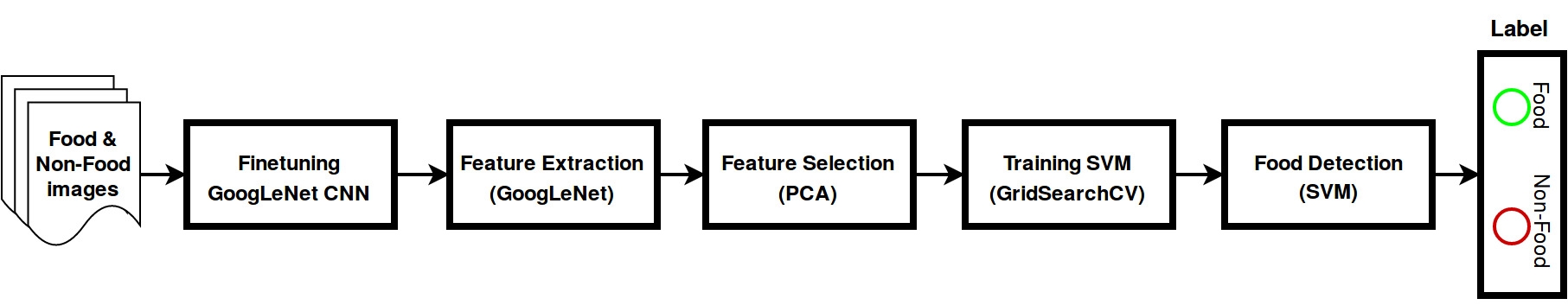}
\caption{\small Method overview for our food detection approach.}
\label{fig:fig3}
\end{figure}

We propose a methodology for food detection, which involves the use of the GoogLeNet model for feature extraction, PCA for feature selection and SVM for classification. In the Fig.\ref{fig:fig3} we showed the pipeline of our food detection approach which will be explained below.

\subsection{GoogLeNet CNN}

The first step in our methodology involve to train the GoogLeNet CNN model. For this, we take the GoogLeNet previously trained on ImageNet \cite{russokovsky2015}, as a base model, and then we 
change the number of classes in the output layer from 1000 to 2, because in our case is a binary classification (food/non-food). After that, GoogLeNet is fine-tuned on the last two layers until the accuracy on training set stop to increase, then we choose the model that give the best accuracy on the validation set. 

Once GoogLeNet was fine-tuned, we use the resulting model as feature extration method. The feature vector for each image is extracted using the layer before the last one, with which it is obtained a vector with 1024 dimensions by image. Then, by mean of the feature vectors obtained from the training set, we calculate a transformation that distribute normally the data throught a gaussian distribution function with zero mean and unit variance. Finally, we normalize the data by applying this transformation to each extracted feature vector.

\subsection{PCA}

The following is to reduce the dimensions of the feature vectors obtained in the previously steps by means of a linear transform named Principal Component Analysis (PCA), which transforms the data to a new coordinate system leaving the greatest variance of the images in the first axes (principal components). We apply PCA on all feature vectors normalized from training set and then the principal components are analysed to select the first dimensions that retain the most discriminant information. To do this, we selected the features based on the Kaiser Criterion \cite{kaiser1960}, which consists of retaining those components with eigenvalues greater than 1. The feature vectors reduced are used during the training of SVM and also during the classification.

\subsection{SVM}

The training of the SVM classifier is make by mean of the GridSearchCV strategy on 3-folds using the feature vectors reduced obtains from the training set images. As in \cite{ragusa2016}, we opt to use the sigmoid function as kernel and we find the best combination of cost (C) and gamma parameters. After, we selected the parameters that prov the best result and use it to train the SVM from scratch with all training set.

\subsection{Classification}
Finally, with SVM trained, the last step is to classify the images. To Summarize the process, for each image we get a feature vector with GoogLeNet fine-tuned, and after, we normalize it (2.1), then the vector dimensions is reduced with PCA (2.2) and finally it is classified with SVM (2.3), obtaining the corresponding class to the image (food/non-food). 

\section{Experiments}

\subsection{Datasets}

In this section, we present the selected datasets from the previous works for the evaluation of the proposed models and contrast the results. Both datasets, FCD and RagusaDS, were selected because they contains a significant amount of images, at least 15,000 images, and also because the access to their images is free.
\subsubsection{FCD}
\begin{figure}
\centering
\includegraphics[width=.95\textwidth]{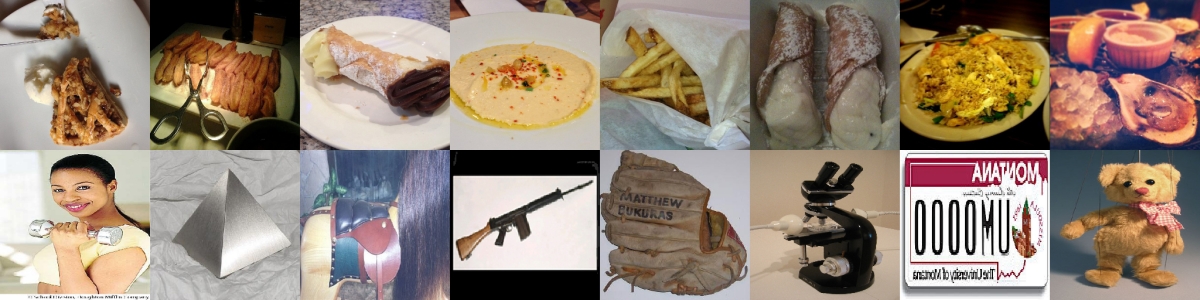}
\caption{\small Example of images contained in the FCD Dataset. Top row shows food images from Food101 and bottom row shows non-food images from Caltech256.}
\label{fig:fig1}
\end{figure}

The dataset was constructed from two public datasets widely used: Food-101 \cite{food101} and Caltech-256 \cite{griffin2007} for food and non-food images (see Fig.\ref{fig:fig1}). Food-101 is a benchmark dataset usually used for the food recognition task; this dataset contains 101 popular food categories downloaded from foodspotting.com, with 1,000 images for each dish. As for Caltech256, it contains 256 categories of objects with a total of 30,607 images, in which each object has a minimum of 80 images. For the construction of FCD, not all images of these datasets were considered, to avoid highly imbalance between images of food and non-food. Therefore, in the case of Food101, 250 images are selected for each category through the analysis of color histogram of images, leaving those with the highest color variance within the same category, obtaining a total of 25,250 food images. The images was selected based on the histogram because to the importance of Color feature in the food images recognition problem. In the case of Caltech 256,  all images are selected except those which contain food-related objects, leading to a total of 28,211 non-food images. To evaluate our approach, we used the 20\% of the images for the testing set, and of the remaining 80\%, a 80\% is used for the training set and a 20\% for validation set.

\subsubsection{RagusaDS}
\begin{figure}
\centering
\includegraphics[width=.95\textwidth]{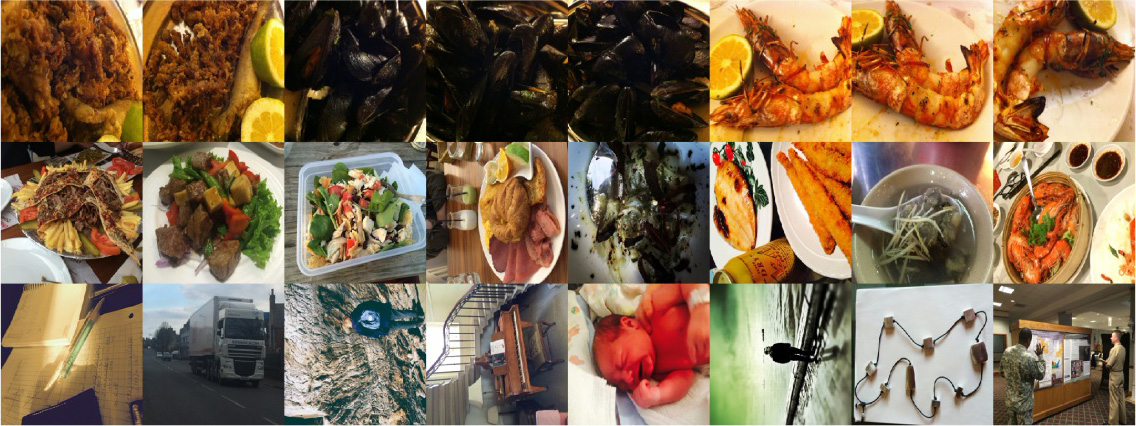}
\caption{\small Example of images contained in the RagusaDS Dataset. Top row shows food images from UNICT-FD889, middle row shows food images from Flickr-Food, and bottom row shows non-food images from Flickr-NonFood.}
\label{fig:fig2}
\end{figure}

The dataset was constructed from three datasets with images acquired in different conditions: UNICT-FD889 \cite{farinella2015ds} and Flickr-Food \cite{farinella2015} for food images and Flickr-NonFood \cite{farinella2015} for images other than meals. UNICT-FD889 is a dataset containing 3,583 images of meals of 889 different dishes acquired from multiple perspectives with the same device in real-world scenarios, where images of meals are acquired from a top view avoiding the presence of other objects. Another part of these dataset is composed by images downloaded from Flickr and manually labelled as being food or non-food images. These datasets, which are called Flickr-Food and Flickr-NonFood, contain 4,805 images of food and 8,005 non-food images, respectively. Differently to UNICT-FD889, these datasets contain images less restricted, and specifically for Flickr-Food the images can contain objects different to the food and can be taken from different points of view. In total, the dataset contains 8,388 images of food and 8,005 of non-food (see Fig.\ref{fig:fig2}). To evaluate our approach, we used for training the 80\% and for validation the 20\% of all images from UNICT-FD889 together with 3,583 images contained in the first half of Flickr-NonFood and for testing all images from Flickr-Food together with 4,422 images contained in the second half of Flickr-NonFood.

\subsection{Experimental Setup}

The  fine-tuning of the CNN models were through Caffe Framework \cite{caffe}, which is one of the most popular Deep Learning framework used in the resolution of challenges in the field of computer vision. In addition, we selected it because of its simplicity to use and especially because it provides multiple CNN models previously trained on the ImageNet dataset, among which is the GoogLeNet. We fine-tune the last two layers of this model by assigning ten times the default learning rate in each.

With respect to the parameters used to train the network, we initialized the learning rate to $1 \times 10^3$, configured the drop learning rate every 5,000 iterations and used a batch size of 32. As for the pre-processing of images used during the training and validation of GoogLeNet, we resized the input images to a size of 256x256, we subtracted the images average from the ImageNet dataset and maintained the original scale. In particular for the training, the simple data augmentation was generated by horizontal mirror and random crops of a 224x224 dimension. As for the validation and evaluation, a center crop is done to the image of 224x224 dimension.

On the other hand, the GoogLeNet is fine-tuned during 10 epochs, in the case of FCD, and during 40 epochs for RagusaDS, using the training set of the respective datasets. The resulting model that prov the best accuracy on the validation set, in each dataset, is used to extract the images features. The feature vector calculate from the image is reduced selecting the principal components obtained by mean of the training set, so, we obtain a feature vector with 186 dimensions on RagusaDS and 206 dimensions on FCD, for each image. As for the values used for C and gamma, during the training of SVM with the GridSearchCV strategy, we defined a range of fourteen values uniformly distribuited on a base-10 logarithmic scale. In the case of C param, we used a range from $1 \times 10^{-4}$ to $1 \times 10^{2}$ and for gamma param from $1 \times 10^{-8}$ to $1 \times 10^{-2}$.

\subsection{Metrics}

We used three metrics to evaluate the performance of our approach, overall Accuracy ($ACC$), True Positive rate ($TPr$) and True Negative rate ($TNr$), which are defined as follows: $ ACC = \frac{TP + TN}{T},$ where $TP$ (True Positive) and $TN$ (False Negative) are the amount of correctly classified images as Food and Non-Food respectively; 
$TPr = \frac{TP}{TP + FN}$, where $FN$ (False Negative) is the amount of misclassified images as Non-Food;
$FNr = \frac{FN}{FP + TN}$,where $FP$ (False Positive) is the amount of images misclassified as food.

\subsection{Results}

In this section, we present the results obtained during the experimentation. In Table \ref{tab:tab1}, the first two rows correspond to state of the art algorithms that gave the best prediction on RagusaDS and FCD datasets, respectively. The last three methods are variations of our proposal, which is based on the GoogLeNet. The results show the ACC, the $TPr$ and $TNr$ obtained when evaluating each method on the FCD and RagusaDS datasets. In the case of the FCD, it can be seen that the model allows obtaining a high precision in the global classification and maintaining a slightly higher precision for $TNr$, which may be due to the small imbalance between food and non-food images from this dataset. On the other hand, for RagusaDS the difference between $TPr$ and $TNr$ is about 7\% better for $TNr$. We believe that this is mainly due to the fact that the food images used during the training are very different from those used for the evaluation and therefore the model is not able to recover enough discriminant information that allows it to generalize over a sample acquired under different conditions.


\begin{table} \caption{Results obtained by models based on CNN on RagusaDS and FCD datasets on the food detection task. All results are reported in \%.}  
\centering
\setlength{\tabcolsep}{0.5em}
\begin{tabular}{  l  c c c | c c c }
    & \multicolumn{3}{c}{RagusaDS} & \multicolumn{3}{c}{FCD} \\ 
     \cline{2-7}
     & \textbf{ACC} & \textbf{TPr} & \textbf{TNr} & \textbf{ACC} & \textbf{TPr}& \textbf{TNr}\\ 
    \hline

    AlexNet + SVM \cite{ragusa2016} &94.86	&\textbf{94.28}	&95.50	&-	&-	&- \\ 

    NIN  \cite{kagaya2015} &-	&-	&-	&96.4	&96	&97 \\ 
    \hline	
    \hline			
    GoogLeNet + PCA-SVM & \textbf{94.97} & 91.57 & \textbf{98.67} & \textbf{99.01} & \textbf{98.85} & 99.15\\ 

    GoogLeNet + SVM &  94.95 &91.53 &\textbf{98.67} &98.96 &\textbf{98.85} &99.06 \\ 

    GoogLeNet &94.66	&91.53	&98.06	&98.87	&98.48	&\textbf{99.22} \\ 
	\hline
    
\end{tabular}
\label{tab:tab1}
\end{table}

GoogLeNet+PCA-SVM is selected for the next experiment because it is the model that achieved the best results on both datasets. Then, we trained it and evaluated its performance using RagusaDS and FCD datasets together, maintaining the same sets of training, validation and test, which we named RagusaDS+FCD. The Table \ref{tab:tab2} shows the results obtained by training of our approach using the training set from RagusaDS+FCD and evaluating it on the test sets from RagusaDS+FCD, RagusaDS and FCD. The results shows that, when is trained the model on RagusaDS+FCD, it improves the classification significantly on RagusaDS although it presents a slight decrease on FCD. We believe that the improvement on RagusaDS is mainly due to an increase in the detection of food-related images, so we deduced that by combining the datasets, during training, the proposed method was able to extract features from various types of food acquired in different conditions, which allowed to have a more robust classifier achieving a better generalization on the test set of the RagusaDS dataset.

\begin{table} \caption{Results obtained when GoogLeNet+PCA-SVM is trained with both datasets together (RagusaDS+FCD) and evaluated separately and jointly.} 
\centering
\setlength{\tabcolsep}{0.5em}
\begin{tabular}{ l c c c  }
    \multicolumn{1}{c}{\textbf{Test dataset}} & \textbf{ACC} & \textbf{TPr} & \textbf{TNr} \\ 
    \hline
    \hline
    \multicolumn{1}{l|}{RagusaDS} & 95.78\%	&93.65\%	&98.10\% \\ 
    \multicolumn{1}{l|}{FCD}   &98.81\%	&98.60\%	&99.01\% \\
    \multicolumn{1}{l|}{RagusaDS+FCD}  &97.41\%	&96.19\%	&98.61\% \\ 
	\hline
    
\end{tabular}
\label{tab:tab2}
\end{table}

\begin{figure}
\centering
\begin{subfigure}
  \centering
  \includegraphics[width=0.48\textwidth]{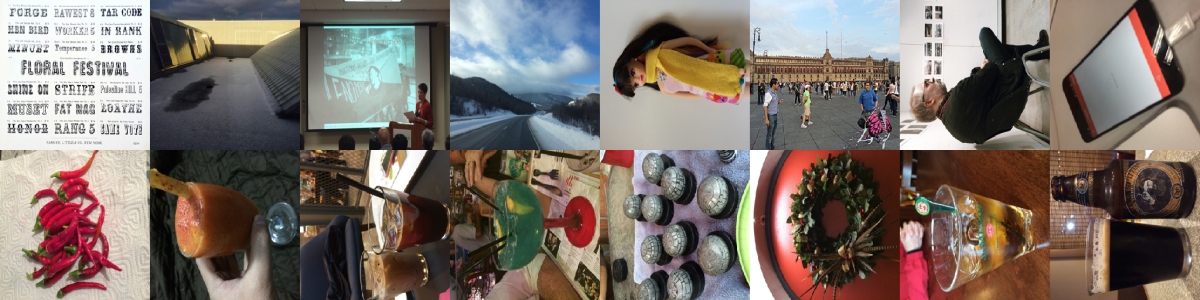}
\end{subfigure}
\centering
\begin{subfigure}
  \centering
  \includegraphics[width=0.005\textwidth]{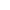}
\end{subfigure}
\begin{subfigure}
  \centering
  \includegraphics[width=0.48\textwidth]{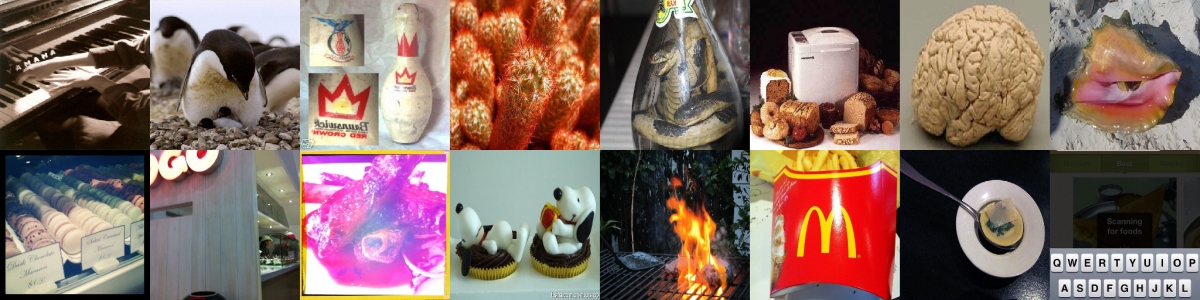}
\end{subfigure}
\vspace*{-.5cm}
\caption{\small FP (top) and FN (bottom) on RagusaDS (left) and FCD (right) datasets.}
\label{fig:fig4}
\end{figure}

The False Positives (FP) and the False Negatives (FN) obtained in both dataset are shown in Fig. \ref{fig:fig4}. 
When analyzing the FP can be observed that in the case of RagusaDS 
most errors occurred in images in which the food was a liquid (drink, coffee, etc). We think that the reason for this is mainly because the train set contains a wide variety of dishes but none of these correspond to the beverages and therefore the classifier does not recognize them as food. In addition, other factors that influence classification are poorly labeled images such as food and also when in the same image there are a lot of dishes. In the case of FCD, there are also some errors caused by wrong labels in both categories. 

\vspace*{-1.0cm}

\section{Conclusions}

In this paper we addressed the food detection problem and proposed a model that use GoogLeNet for feature extraction, PCA for feature selection and SVM for classication. From the results obtained we observed that the best accuracy is achieved in both datasets with our proposed approach. Specifically, the improvement in the overall accuracy is more than 2\% on FCD and about 1\% for RagusaDS, when both datasets are combined for training and evaluated on the respective datasets. In addition, the overall accuracy when combining both datasets is 97.41\%. As a conclusion, we explored the problem of food detection comparing the last works in the literature and our proposed approach provides an improvement on the state of art with respect to two public datasets. Moreover, models based on GoogLeNet, independently of the settings, gave the highest accuracy on the food detection problem. For future work, we will evaluate the performance of CNN-based models on larger datasets containing a much wider range of dishes and beverages such as food images and diversity of environments for non-food images. 

\section*{Acknowledgement}
This work was partially funded by TIN2015-66951-C2, SGR 1219, CERCA, \textit{ICREA Academia'2014} and Grant 20141510 (Marat\'{o} TV3). The funders had no role in the study design, data collection, analysis, and preparation of the manuscript. We acknowledge Nvidia Corporation for the donation of a Titan X GPU.  Eduardo Aguilar acknowledges financial support of CONICYT Becas Chile.


\end{document}